# In-Vehicle Object Detection in the Wild for Driverless Vehicles

Ranjith Dinakaran and Li Zhang

*Computer and Information Sciences, Northumbria University, Newcastle upon Tyne, UK*
*E-mail*: li.zhang@northumbria.ac.uk

Richard Jiang

*Computing and Communication, Lancaster University, Lancaster, UK*
*E-mail*: r.jiang2@lancaster.ac.uk

In-vehicle human object identification plays an important role in vision-based automated vehicle driving systems while objects such as pedestrians and vehicles on roads or streets are the primary targets to protect from driverless vehicles. A challenge is the difficulty to detect objects in moving under the wild conditions, while illumination and image quality could drastically vary. In this work, to address this challenge, we exploit Deep Convolutional Generative Adversarial Networks (DCGANs) with Single Shot Detector (SSD) to handle with the wild conditions. In our work, a GAN was trained with low-quality images to handle with the challenges arising from the wild conditions in smart cities, while a cascaded SSD is employed as the object detector to perform with the GAN. We used tested our approach under wild conditions using taxi driver videos on London street in both daylight and night times, and the tests from in-vehicle videos demonstrate that this strategy can drastically achieve a better detection rate under the wild conditions.

## 1. Introduction

Vision-based object tracking and detection plays an important role in the recently emerging self-driving vehicle systems [1-6]. Particularly following the accident in Uber's self-driving test that killed a pedestrian [1], it has been raised as a major barrier on how to secure the self-driving vehicles under the wild conditions. If a self-driving system cannot overcome this challenge, driverless vehicles will not be a bliss but a curse to smart cities with road killers running everywhere.

To tackle with in-vehicle object detection under wild conditions, in this work, we aim to leverage Deep Convolutional Generative Adversarial Network (DCGAN) [7-9], a powerful method to enrich the data coverage and consequently improve the performance of deep neural networks (DNNs). We propose to combine DCGAN with Single Shot Detector (SSD) [8-9] for in-vehicle object detection in the wild. Such a DCGAN-SSD framework has been shown very efficient in image-based object detection. In this work, we will particularly leverage it for night-vision vehicle guidance. While our application is targeted at the in-vehicle object detection for self-driving, to guarantee the speed of object detection, SSD as the object detector is favored due to its performance. In comparison [8-9] with RCNN, Fast RCNN and Faster RCNN, SSD is several orders of magnitude faster in term of computing time, making it a suitable candidate for the real-time applications such as self-driving.

In our work, we investigate the feasibility of applying GANs to datasets for in-vehicle object detection. In our practical, we train our GAN with a set of visual objects from CIFAR dataset as the $1^{st}$ step. The discriminator will cope with the task of discriminating the real image containing the visual objects provided in the dataset from the generated image, with the generated background pixels between the visual objects. The main motivation of this work starts from the training data for object detection [7-10] and we will focus on GANs especially in wide amount of data technique to improve an existing object detection task.

## 2. Preliminary

### 2.1. *Generative Adversary Networks*

The Generative Adversarial Networks (GANs) [7] is an outline for learning generative models. GANs have been applied to image super resolution and improve the image quality [9-10]. It has also been attempted to accommodate GANs on the object detection task to address the small-scale problem by generating super-resolved representations for small objects [8-9]. Recently several attempts have been made to improve image generation using generative models. The most popular generative model approaches are Generative Adversarial Networks (GANs) [7], Variational Autoencoders (VAEs), and Autoregressive models. Their variants, e.g. conditional GANs, reciprocative conditional GANs, and Deep Convolutional GANs (DC-GANs), etc..

Radford et al. [10] use a Conv-Deconv GAN architecture to learn good image representation for several image synthesis tasks. Denton et al. [11] use a Laplacian pyramid of generators and discriminators to synthesize multi-scale high resolution images. Reed et al. [12] use a DC-GAN conditioned on text features encoded by a hybrid character-level convolutional RNN. Perarnau et al. [13] use an encoder with a conditional GAN (cGAN), to inverse the mapping of a cGAN for complex image editing, calling the result Invertible cGANs. Makhzani et al [14] and Larsen et. al. [15] combine a VAE and GAN to improve the realism of the generated images.

### 2.2. *Deep Convolutional GAN*

Let y be the original input image, and y' represent the generated image from the generative model give a noise vector sampled from x ~ uniform (0.1). Let D(z) and D(z') be the output of the discriminator network for the two images and represent the regressed confidence that the input is from the number of real images. Intuitively, we seek to minimize D(z') and maximize D(z). The value function id D that we can write is now:

$$\log(D(x)) + \log(1-D(z')) \qquad (1)$$

The complete minimax value function V for G and D now becomes as defined by Goodfellow [7]:

$$\min \max V(D, G) = E_z(\log D(z)) + E_{z'}(\log(1-D(z'))) \qquad (2)$$

The challenging task is image decryption. Whereas G can choose any point in the generator multiple images to consider a valid image, in the decryption task, we first consider G with a set of predetermined pixels based on ground truth, these pixels can be a random pixel. The experiment becomes very challenging because now a very small number of multiple images will produce an accurate image given to the existing images. This is typically achieving by backpropagation through a multi-component reconstruction loss.

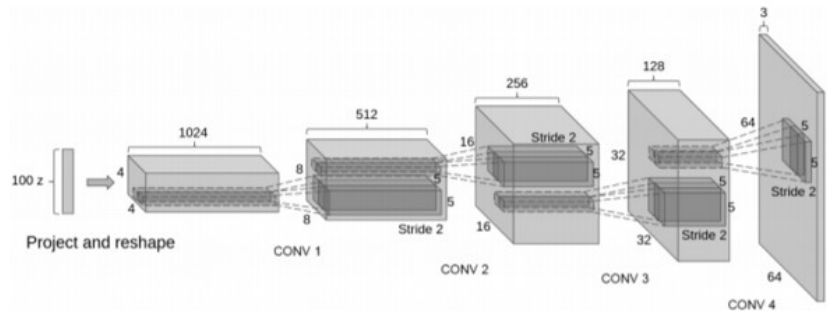

Fig 1. The DCGAN architecture.

Fig 1 shows the architecture used for the generative portion of this system. The noise vector x is expanded through a series of deconvolutional layers. Our Network input is 19x19 dimension where the labelled visual objects are retained in original configuration, and the left-over pixels are sampled and resampled from noise distribution heuristically Gaussian approach. The image experiences a series of convolutional layers while its retaining the same dimensions. We found that in this resolution GANs generator part will not congregate nor begin to learn which is appropriate in data setting[8-9].

## 3. Using DCGAN-SSD for In-Vehicle Object Detection

### 3.1. *Using DC-GAN to improve the input frames*

In our proposed strategy, we aim to improve the video frames from the wild using DC-GAN. Fig.2 shows a test example. Fig.2-a) shows the raw image obtained from the video, with a low pixel ratio of 320x100, and with the size of the 27kb. Fig.2-b) is the produced high resolution image where the pixels and the ratio of the image been enhanced, after enhancement 1920x1080, after enhancement the size of the image also been enhanced to 451kb.

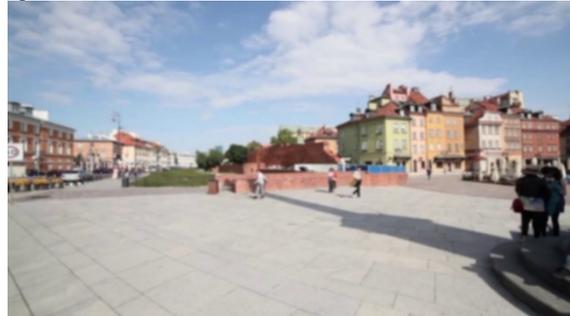

a) Before our trained DC-GAN

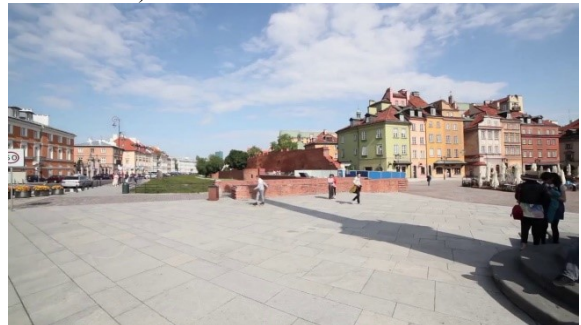

b) After our DC-GAN

Fig2. The generated high-res images from low-res ones.

## 3.2. DCGAN-enhanced SSD-based object detector

Object detection is considered as a major challenge of the image classification task, where the main goal is to classify and localize every object from input image. The object detection problem is considered as a major challenge in computer vision and, made some progress in recent years because of advanced machine learning tools like deep learning and GANs [7]. There are two main region proposal methods: You only look once (YOLO) and SSD. When we cascade DCGAN+SSD the main advantage is, the combination reduces the scale factors in the images, and also with SSD we add convolutional feature layers to the end of curtailed base network. These layers in GAN and SSD reduce the size progressively at different scales. The model of making predictions varies for each feature layer that operates in SSD.

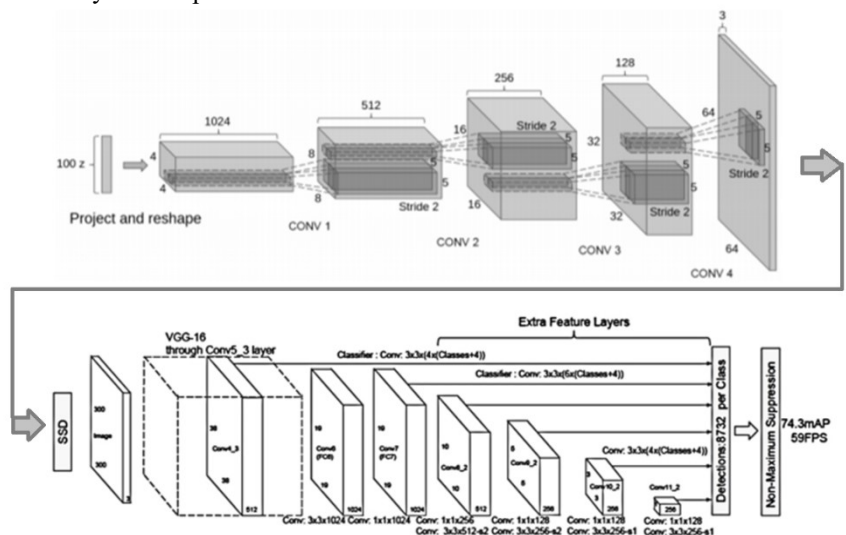

Fig 3. The Architecture of DCGAN + SSD [8, 9].

The most communal technique for validating the eminence of unsupervised illustration learning algorithms is to apply them as a feature extractor on supervised datasets and validate the performance of linear models tailored on top of these features as shown in Fig 1 the convolutional layers are leveraged to resize the images and there is no room for enhancement of images or to fill the feature space. However the Fig.3 demonstrates the changes in the convolutional layers and these layers in the GAN is built to enhance the image quality under any size, so that the feature space is filled with feature maps for better detection quality.

We use the pretrained SSD network to save the computational time, and then use the discriminator's convolutional features from all layers, max-pooling each layers representation to produce a 4 × 4 spatial grid for 32x32 size images. These features are then flattened and concatenated to form a dimensional vector and a regularized linear classifier is trained on top of them. This achieves better accuracy than by using SSD on its own.

## 4. Validation on the in-the-wild in-vehicle videos

In our experiments, DCGAN was implemented in two steps. First, we implemented the DCGAN codes based on PyTorch and Tensorflow, as illustrated in [8-9]. We trained our DCGAN on the CIFAR, Caltech, KITTI dataset, with all image size of 32×32, along with the batch size of 72, and 25 epochs, across a total of 60,000 images. We then used CIFAR, Caltech, and KITTI dataset to train our DCGAN+SSD detector. We aim to validate if the proposed GAN + SSD architecture outperforms the single SSD, particularly on distant object or pedestrian detection. Following this, we ran our trained DCGAN+SSD detector on in-the-wild videos taken from in-vehicle cameras, and compare their performance against SSD-only detector.

Fig.4 shows our test results on these videos taken from in-vehicle cameras. The videos consist of daytime on-the-street videos and night-vision videos. From the test samples, we can clearly see the test results are overwhelmingly better when DCGAN is attached to improve the performance of SSD. Particularly, the DCGAN+SSD detector can even detect those pedestrians in the dark shadowing regions, which is critical and very challenging for vision-based driverless vehicle systems.

As it was reported, Uber suspended its driverless vehicle project due to the accident of its driverless vehicle in the night, while a pedestrian crossing the road in the night was ignored by its object detection system and tragically, was hit by the Uber's driverless vehicle (in test). Our experiment, hopefully, provides a reliable object detection in the night and hence, could be complementarily useful for a vision-based driverless system.

Notably, the SSD detector performance lags in the scale factor, which is compromised by DCGAN in our experiment, where the DCGAN can improvise the scale factor in SSD by providing the detector with super resolution images, it does result in a larger total feature vector size due to the highest layers given for feature vectors of 4 × 4 spatial locations. Further improvements could be made by fine-tuning the discriminator's representations, but we leave this for future work. We will consider how to make the detector robust to noises [16], and include 3D CNN as an alternative solution for object detection in videos [17-21].

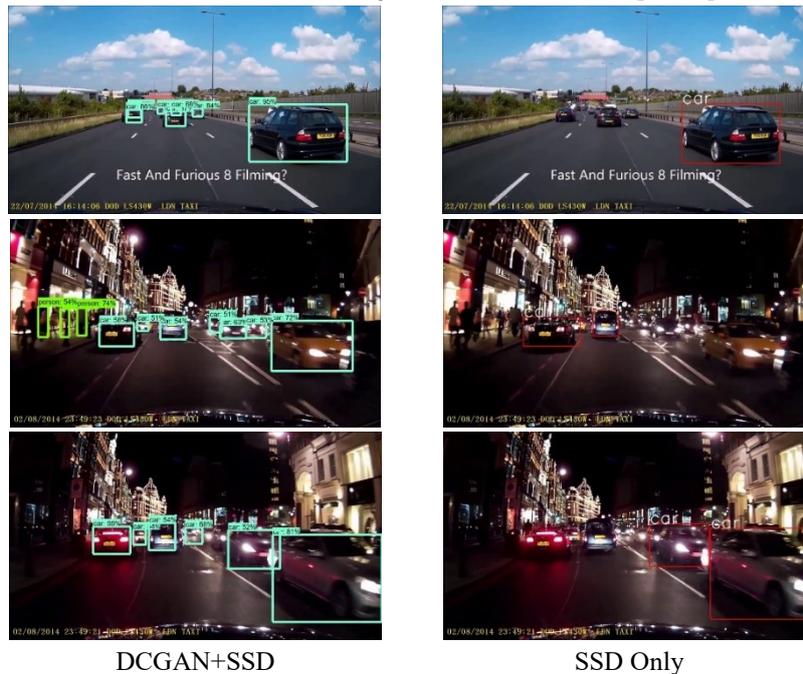

        DCGAN+SSD                            SSD Only

Fig.4 Experimental results on wild videos. More details on Youtube videos: https://youtu.be/jwTnXDFPUMc, https://youtu.be/aQ202pcZWXE.

## 5. Conclusions

While in-vehicle human object identification plays an important role in vision-based automated vehicle driving systems, pedestrians on roads or streets are the primary targets to protect from driverless vehicles in smart cities. A challenge is the difficulty to detect objects in moving under the wild conditions, and illumination and image quality could drastically vary. In our work, to address this challenge, we examined the feasibility of applying DCGANs with Single Shot Detector (SSD) to handle with the wild conditions. The GAN was trained with low-quality images to handle with the challenges arising from the wild conditions in smart cities, and a cascaded SSD is employed as the object detector to perform with the GAN. We used Canadian Institute for Advanced Research (CIFAR),

Caltech, KITTI data set for training. Further in-vehicle tests on the London streets demonstrate that this strategy can drastically achieve a better detection rate under the wild conditions.